\documentclass[10pt,twocolumn,letterpaper]{article}
\pdfoutput=1
\usepackage{cvpr}
\usepackage{times}
\usepackage{epsfig}
\usepackage{graphicx}
\usepackage{amsmath}
\usepackage{amssymb}

\usepackage{bm}
\usepackage{microtype}
\usepackage{url}
\usepackage{multirow}
\usepackage{subfig}

\usepackage{color}
\newcommand{\new}[1]{{\textcolor{black}{#1}}}

\newlength{\eqs}
\usepackage[pagebackref=true,breaklinks=true,letterpaper=true,colorlinks,bookmarks=false]{hyperref}

\cvprfinalcopy 
\vspace{-0.9cm}
\def\cvprPaperID{336} 

\ifcvprfinal\fi

\begin{document}

\title{WarpNet: Weakly Supervised Matching for Single-view Reconstruction}

\author{Angjoo Kanazawa\enskip\qquad David W. Jacobs\\
University of Maryland, College Park\\
\and
Manmohan Chandraker\\
NEC Labs America\\
}

\maketitle

\begin{abstract}
We present an approach to matching images of objects in fine-grained datasets
without using part annotations, with an application to the challenging problem
of \new{weakly supervised} single-view reconstruction. This is in contrast to prior works that require part annotations, since matching objects across class and pose variations is challenging with appearance features alone. We overcome this challenge through a novel deep learning architecture, {\em{WarpNet}}, that aligns an object in one image with a different object in another. We exploit the structure of the fine-grained dataset to create artificial data for training this network in an unsupervised-discriminative learning approach. The output of the network acts as a spatial prior that allows generalization at test time to match real images across variations in appearance, viewpoint and articulation. On the CUB-200-2011 dataset of bird categories, we improve the AP over an appearance-only network by $13.6\%$. We further demonstrate that our WarpNet matches, together with the structure of fine-grained datasets, allow single-view reconstructions with quality comparable to using annotated point correspondences.

\end{abstract}

\vspace{-0.2cm}
\section{Introduction}

\begin{figure}
\begin{center}
\includegraphics[width=\linewidth]{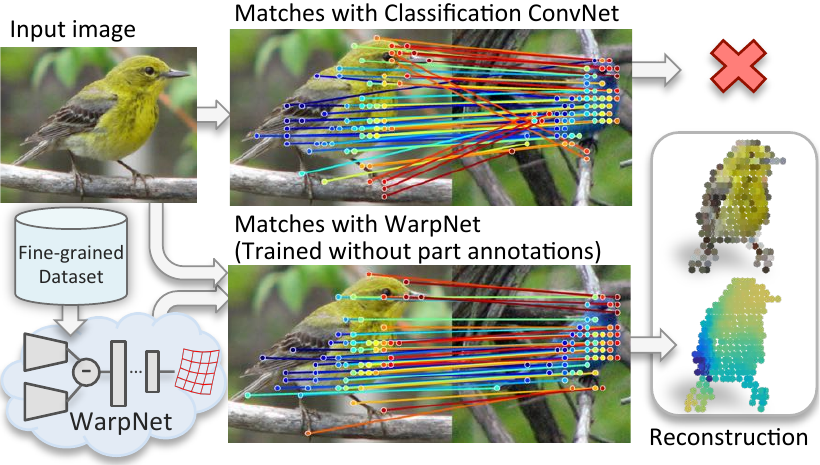}
\end{center}
\vspace{-0.5cm}
\caption{\small Given a single image of an object, we propose a novel deep learning framework for obtaining keypoint matches to other objects in a fine-grained dataset, without using any part annotations. The output of our network is used as spatial prior for matching across variations in appearance, pose and articulation (bottom), which is not possible with appearance features alone (top). Our match quality is high enough to be propagated across images to be used for 
\new{single-view reconstruction without using any manually annotated keypoints}  (right).
}
\label{fig:open}
\vspace{-0.3cm}
\end{figure}

\begin{figure*}
\begin{center}
\includegraphics[width=\linewidth]{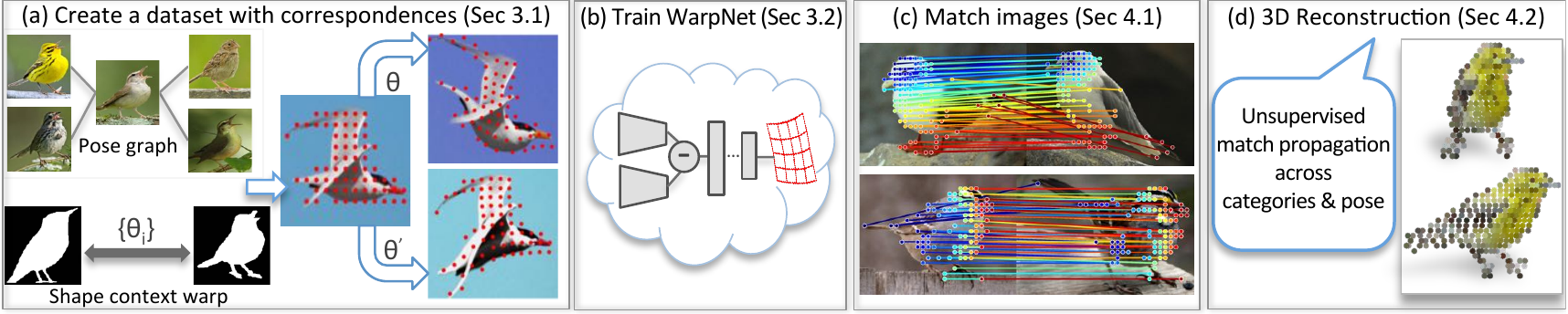}
\end{center}
\vspace{-0.5cm}
\caption{\small Overview of our framework. (a) Lacking part annotations, we
  exploit the fine-grained dataset to create artificial correspondences. (b)
  These are used to train our novel deep learning \new{architecture} 
  that learns to warp one object into another. (c) The output of the network is used as a spatial prior to match across appearance and shape variations. (d) Our high-quality matches can be propagated across the dataset. We use the WarpNet output and the structure of fine-grained categories to perform single-view reconstruction without part annotations.}
\label{fig:overview}
\vspace{-0.3cm}
\end{figure*}

Reconstructing an object \new{from} a single image is a significant challenge, that can be tackled by matching keypoints to other instances in a fine-grained dataset. However, such datasets exhibit large intra-class shape variations or inter-class appearance variations, which cannot be handled by traditional features such as SIFT \cite{lowe2004}.
Recently, methods have been proposed to match instances across categories,
relying on supervision in the form of part (keypoint) annotations
\cite{VVN,Kar,Vincente} or 3D CAD models \cite{Bao,Cashman} to augment
appearance information with shape priors. Such annotations are labor-intensive,
thus, too sparse for reconstruction and not scalable. Further, it can be quite difficult 
to obtain human-labeled annotations for parts that are not nameable.
In contrast, this paper presents a framework to match images of objects with some degree of non-rigidity and articulation, across category and pose variations, without requiring supervised annotations. We then present an approach to the challenging novel problem of unsupervised single-view object reconstruction.

\begin{figure}[!!t]
\begin{center}
\includegraphics[width=0.85\linewidth]{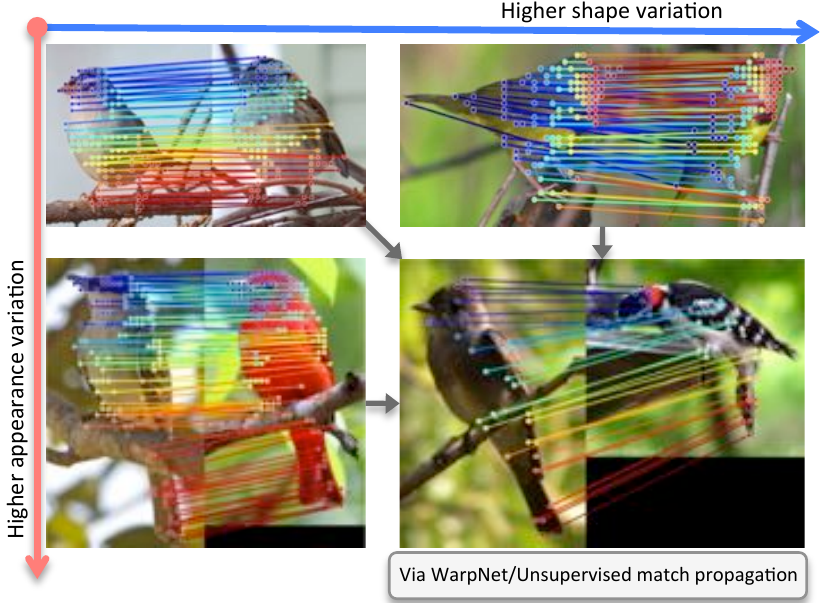}
\end{center}
\vspace{-0.6cm}
\caption{\small Intuition for matching in fine-grained datasets without supervised point annotations. Matching within a category exploits appearance similarity, while matching instances across related categories is possible through global shape similarity. By propagation, one may match across variations in both appearance and shape.}
\label{fig:intuition}
\vspace{-0.4cm}
\end{figure}

We postulate that the structure of fine-grained datasets, combined with the power of convolutional neural networks (CNNs), allows matching instances of different categories without supervised annotation. Fine-grained datasets for objects such as birds can be analyzed along two dimensions -- appearance and shape. Instances within the same category that are imaged in different poses can be matched by appearance similarity, while instances with similar pose or viewpoint from different categories can be matched through similarity in global shape. Instances with both appearance and shape variations may then be matched by propagation (Fig. \ref{fig:intuition}). In Section \ref{sec:learning}, we demonstrate a practical realization of this intuition by introducing a deep learning architecture, {\em{WarpNet}}, that learns to warp points on one object into corresponding ones on another (from a possibly different category or pose), without requiring supervised annotations.

WarpNet is a Siamese network that accepts two images as input (Section \ref{sec:architecture}). To overcome the absence of annotated keypoints, our training presents an image and a warped version related by a known thin-plate spline (TPS) transformation, which yields artificial correspondences. 
We assume the object bounding box and foreground segmentation are known, which
can be obtained through state-of-the-art segmentation \cite{Chen} or
co-segmentation methods \cite{Krause}. \new{We experiment using both ground truth and co-segmentation outputs.}
 In Section \ref{sec:making}, we exploit neighborhood relationships within the dataset through the pose graph of Krause \etal \cite{Krause} to compute exemplar TPS transformations between silhouettes, from which our artificial transformations are sampled. 
A point transformer layer inspired by \cite{Jaderberg} is used to compute the warp that aligns keypoints without supervision, which provides a spatial prior for matching (Section \ref{sec:matching}).
We show that WarpNet generalizes well to match real images with distinct shapes and appearances at test time. In particular, it achieves matching accuracy over $13.6\%$ higher than a baseline ILSVRC CNN \cite{Chatfield}.

Establishing matches between a given instance and other objects in the dataset
opens the door to a novel problem -- \new{weakly supervised} reconstruction in fine-grained datasets. Several sub-problems must be solved to achieve this goal, such as match propagation and image subset selection. Prior works such as \cite{VVN,Vincente} approach these sub-problems, but the absence of supervised annotations poses new challenges. In Section \ref{sec:reconstruction}, we suggest ways to overcome them through the use of matches from our WarpNet, the pose graph and heuristics that exploit the structure of fine-grained datasets. We demonstrate reconstructions that are nearly as good as those obtained using supervised annotations and better than those from appearance-only CNNs or unsupervised baselines such as deformable spatial pyramids \cite{Jaechul}.

To summarize, our key contributions are:
\vspace{-0.15cm}
\begin{itemize}
\setlength\itemsep{-0.1cm}
\item A novel deep learning \new{architecture}
  , WarpNet, that predicts a warp for establishing correspondences between two input images across category and pose variations.
\item A novel exemplar-driven mechanism to train WarpNet without requiring supervised keypoint annotations.
\item An approach to unsupervised single-view object reconstruction that exploits the structure of the fine-grained dataset to yield reconstructions of birds nearly on par with the method that uses supervised part annotations.
\end{itemize}

\vspace{-0.3cm}
\section{Related Work}

\vspace{-0.2cm}
\paragraph{Supervised matching}
Several recent approaches use deep learning to learn a similarity metric between image patches in a supervised manner \cite{Xufeng,zagoruyko}. These works focus on matching images of the same instance (for example, the Statue of Liberty \cite{UBCdata}) from various viewpoints, while we match deformable objects of different instances exhibiting a wide variety of appearances. Our task requires semantic understanding of object shape, beyond just local appearance. A CNN framework to predict dense optical flow on general scenes is proposed by \cite{FlowNet}, but in a supervised manner.

Matching or keypoint localization may be improved by augmenting appearance similarity with spatial priors. Supervised methods often use a dataset with labeled parts to obtain a non-parametric prior on keypoint locations \cite{coe,jlong,Tulsiani}. These priors may be learned from data \cite{tompson}, but require supervised part annotations during training. Such annotation is laborious and consequently available only for a few nameable parts, which might be too sparse for reconstruction.

\vspace{-0.4cm}
\paragraph{Unsupervised matching}
Also related to our approach are methods that use unsupervised spatial priors for dense matching \cite{Jaechul,sift-flow}. Unlike our work, these methods are purely geometric and do not learn category-specific semantic properties. Recently, \cite{FlowWeb} proposes an unsupervised approach for dense alignment of image sets. But while their focus is global consistency, our emphasis is on pairwise matching through the WarpNet framework (for which they use flow). Thus, our contribution is complementary and may be used by their framework. We evaluate quantitatively on deformable bird categories, while they use rigid categories on PASCAL.

\vspace{-0.4cm}
\paragraph{Single-view reconstruction}
A new challenge in computer vision is to reconstruct a target object
from a single image, using an image collection of similar objects. The
seminal work of \cite{Vincente} demonstrates the possibility of a solution, but relies on ground truth part annotations to establish correspondences. The subsequent works of \cite{Kar,VVN} take a step further in using part annotations only during training. In contrast, we do not require part annotations at either train or test time.

\vspace{-0.4cm}
\paragraph{CNNs for learning transformations}
Similar to the recent work of \cite{Pulkit}, we use a Siamese network to predict transformations. The key difference is that predicting the ego-motion transformation in \cite{Pulkit} is a pretext for feature learning, while we directly use the predicted transformation as well as its appearance features for matching. Further, they require ground truth transformation parameters in order to train their network, while we use the structure of the fine-grained dataset to generate artificial correspondences and implicitly optimize the parameters.
Finally, rigid transformations in \cite{Pulkit} are discretized in bins and the task is posed as classification, while our network outputs continuous thin-plate spline transformation parameters with a matching objective.

Our architecture is inspired by the recent spatial transformer network of Jaderberg \etal \cite{Jaderberg}, which introduces a deep learning module to predict a spatial transformation. This acts as an attention mechanism driven by a classification objective.
We extend the idea further to predict a warping function that aligns two object instances in an unsupervised manner. Our approach is in line with the recent work of \cite{Dosovitskiy}, which demonstrates that CNNs can be trained without supervised labels by treating an image patch and its transformed versions as a ``surrogate'' class. However similar to \cite{Pulkit}, the unsupervised training objective of classifying the surrogate class is geared towards learning good features, while we show that the output of our network trained by an artificial dataset actually generalizes to matching real image pairs.

\vspace{-0.1cm}
\section{Learning without Part Annotations}
\label{sec:learning}
\vspace{-0.2cm}

We present a deep learning framework, {\emph{WarpNet}}, that learns the correspondence from one image to another without requiring part annotations. Given two images $I_1$ and $I_2$, our network outputs a function that takes points in $I_1$ to points in $I_2$. We parameterize this function as a thin-plate spline (TPS) transformation since it can capture shape deformations well \cite{BelongieSC}. Inspired by Dosovitskiy \etal \cite{Dosovitskiy}, we generate artificial correspondences by applying known transformations to an image. However, our approach is distinct in using the structure afforded by fine-grained datasets and dealing with non-rigidity and articulations. Our network generalizes well to instances of different categories at test time and we use its output as a spatial prior in computing a match between two objects. Figure \ref{fig:overview} gives an overview of our approach. We discuss each step in detail below.

\subsection{Generating Unsupervised Correspondences}
\label{sec:making}
\vspace{-0.1cm}

\begin{figure}[!!t]
\begin{center}
\includegraphics[width=0.9\linewidth]{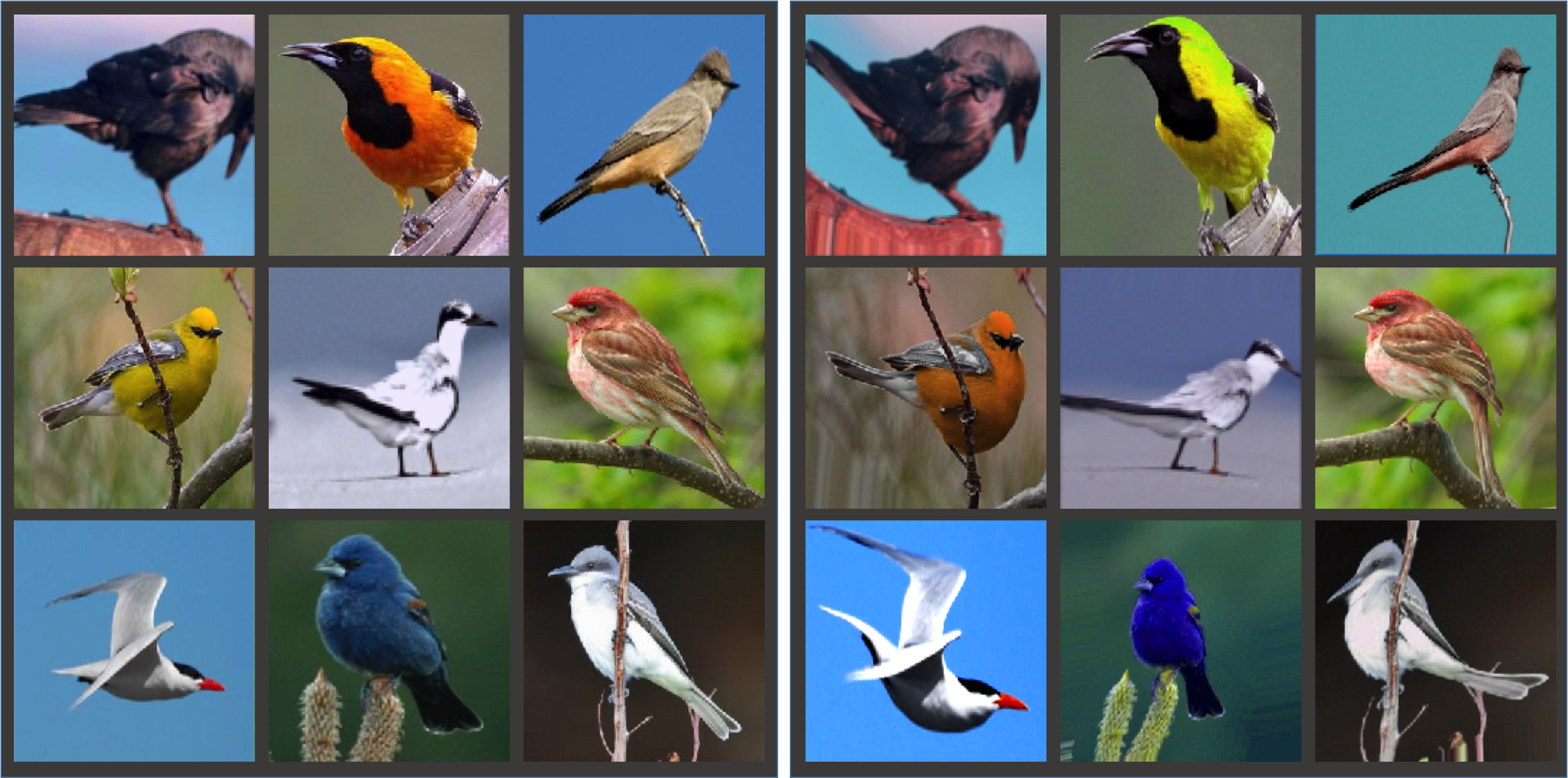}
\end{center}
\vspace{-0.5cm}
\caption{\small Sample exemplar-TPS warped images used for training our WarpNet. Left: original images, right: artificial versions made by applying exemplar TPS warp + chromatic transformation. Notice changes in shape and articulations at the head and the tail. }
\label{fig:jitter}
\vspace{-0.3cm}
\end{figure}

Since we do not have annotated point correspondences, we create artificial ones by applying random spatial and chromatic transformations to images. The key requirement is that the spatial transformations applied are complex enough to learn meaningful correspondences, while producing transformed images that are reflective of actual image pairs to match at test time. For instance, affine transformations are not expressive enough to capture non-rigid deformations and articulations in birds. Instead, we use TPS transformations and exploit the fine-grained dataset to generate exemplar warps that span a realistic range of transformations.

We use the pose graph of Krause \etal \cite{Krause}, whose edge weights are determined by the cosine distance of the fourth-layer of a pre-trained ILSVRC CNN, which captures abstract concepts such as class-independent shape. 
We compute shape context TPS warps \cite{BelongieSC} between the silhouettes of images that are within 3 nearest-neighbors apart on the pose graph.
We sort the TPS warps using the mean of their bending and affine energy, retaining only those between the 50th and 90th percentiles to avoid warps that are too trivial or too drastic. We create $m$ transformed versions of every image by sampling from this set of TPS warps. We sample $n$ points uniformly on the foreground, which we use as correspondences. Figure \ref{fig:jitter} shows the effect of transformations sampled from the exemplar-TPS warps. The images on the left are the originals and the ones on the right are transformed versions. Notice how the transformation induces changes in shape and articulations around the head and the tail, which validates the utility of our exemplar TPS warps.

\subsection{WarpNet Architecture}
\label{sec:architecture}
\vspace{-0.1cm}

\begin{figure}
\begin{center}
\includegraphics[width=\linewidth]{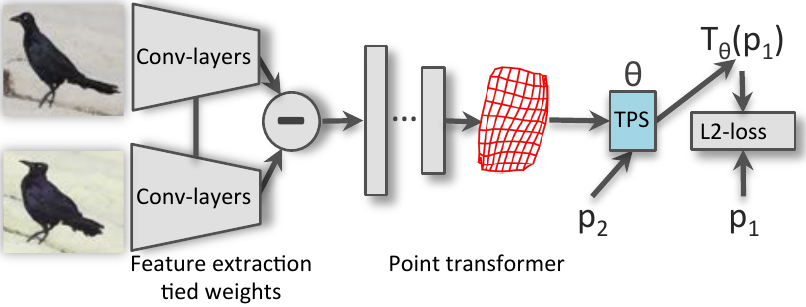}
\end{center}
\vspace{-0.4cm}
\caption{\small WarpNet architecture. Visual features are extracted from two input images using a Siamese CNN. They are combined to predict a deformed grid that parameterizes a TPS transformation. The network objective is to minimize the distance between corresponding points $\bm{p}_1$ and $\bm{p}_2$ of the image pair after applying the predicted transformation to $\bm{p}_2$. }
\label{fig:network}
\vspace{-0.4cm}
\end{figure}

Our proposed WarpNet is a Siamese network \cite{chopra} that takes two images related by an exemplar TPS transformation, $I_1$ and $I_2$, along with the corresponding $n$ keypoint locations, as inputs during training (at test time, the input consists only of two images from possibly different categories and poses that must be matched). The main objective of WarpNet is to compute a function that warps points $\bm{p}_2$ in $I_2$ to image coordinates in $I_1$, such that after warping the L2 distance to the corresponding points $\bm{p}_1$ in $I_1$ is minimized. Figure \ref{fig:network} illustrates the architecture of WarpNet.

First, the input images are passed through convolution layers with tied weights. The extracted features are then combined by element-wise subtraction of the feature maps. We subtract rather than concatenate the feature maps along the channels, since concatenation significantly increases the number of parameters in the network making it unstable to train. The combined feature maps are passed through a point transformer, similar to \cite{Jaderberg}, which regresses on the $(x,y)$ coordinates of a deformed $K\times K$ grid. The output grid, normalized to a range of $[-1, 1]\times [-1,1]$, acts as the control points for computing a grid-based TPS transformation from $I_2$ to $I_1$. This involves solving a system of linear equations, handled by the TPS layer. Please see the supplementary materials for details. The predicted TPS transformation is applied to the keypoints of $I_2$ generating the transformed version $T_\theta(\bm{p}_2)$, which finally gets sent to the L2 loss layer along with $\bm{p}_1$. Since every step consist of linear operations, the whole network can be trained with backpropagation.

We implicitly train the warp parameters in terms of distance between
corresponding points rather than direct supervision against the TPS warp
coefficients. This provides a natural distance between warps, where we can train the network without knowing the exact transformation parameters used. 

\begin{figure}
\begin{center}
\includegraphics[width=0.85\linewidth]{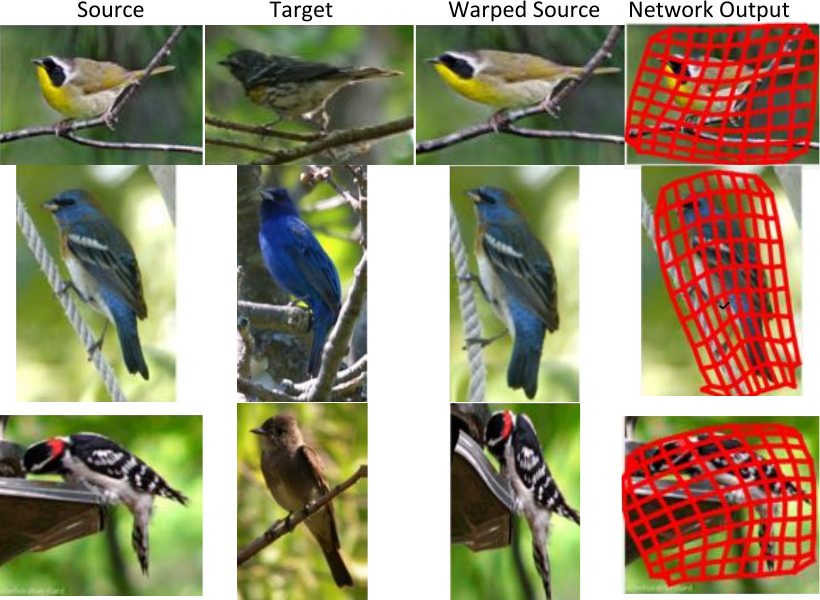}
\end{center}
\vspace{-0.4cm}
\caption{\small Visualizations of the network output. WarpNet takes two images, source and target, as inputs and produces a 10x10 deformed lattice (last column) that defines a TPS warp from target to source. The third column shows the warped source image according to the network output. Notice how the network accounts for articulations at the tail and the head as well as differences in shape of the birds. WarpNet is trained in an unsupervised manner and none of these images were seen by the network during training.}
\label{fig:networkout}
\vspace{-0.4cm}
\end{figure}

Figure \ref{fig:networkout} illustrates the output of the trained network given two real images as input, denoted source and target. Despite the fact that the network has never seen objects of different instances, it is able to compute warps between the two objects. Note that WarpNet accounts for variations in shape (fat to skinny, small to large birds), articulation (such as the orientation of the head or the tail) and appearance.  

\vspace{-0.2cm}
\section{Matching and Reconstruction}
\label{sec:matching}

\vspace{-0.1cm}
\subsection{Matching with WarpNet}
\label{sec:matchcost}
\begin{figure}[!!t]
\begin{center}
\includegraphics[width=0.85\linewidth]{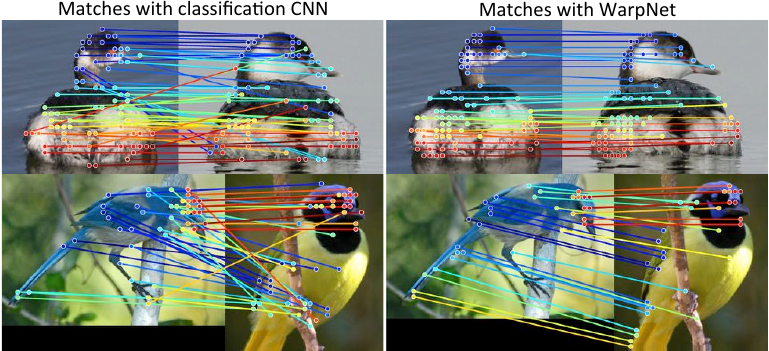}
\end{center}
\vspace{-0.5cm}
\caption{\small Sample matches obtained by ILSVRC trained CNN versus WarpNet, where WarpNet's relative robustness to variations in appearance, pose and articulation may be noted.}
\label{fig:qual_match}
\vspace{-0.5cm}
\end{figure}

\vspace{-0.2cm}
Given two images $I_i$ and $I_j$, a match for a point $u_i$ in $I_i$ is the most similar point $v_j$ in $I_j$ using the similarity score consisting of an appearance term and a spatial term:
\vspace{\eqs}
\begin{equation}
s(u_i, v_j) = \exp \left( \frac{-d_f(u_i, v_j)}{\sigma_f} \right) + \lambda \exp \left( \frac{-d_w(u_i, v_j)}{\sigma_w} \right),
\label{eq:score}
\vspace{\eqs}
\end{equation}
where $d_f(u,v)$ is the L2 distance of appearance features extracted at $u_i$ and $v_j$, while $d_w$ is a symmetric spatial prior:
\vspace{\eqs}
\begin{equation}
d_w(u,v) = \frac{1}{2} (||\bm{x}_i^u -T_{\theta_{ij}}(\bm{x}_j^v)|| +
||\bm{x}_j^v - T_{\theta_{ji}}(\bm{x}_i^u)||).  
\label{eq:warpcost}
\vspace{\eqs}
\end{equation}
We use WarpNet to compute $T_{\theta_{\cdot, \cdot}}$ in both directions.
The matches are then ranked by the ratio-test strategy
\cite{lowe2004}, which allows discarding points in $I_i$ that are similar to many other points in $I_j$. Since the keypoints are extracted densely on the foreground, we compute the similarity score ratio between the first and second nearest neighbors that are at least 10 pixels away. Figure \ref{fig:qual_match} shows a few qualitative matching results comparing the baseline CNN and WarpNet.

\subsection{Single-View Object Reconstruction}
\label{sec:reconstruction}
\vspace{-0.2cm}

Obtaining good matches is a critical first step towards 3D reconstruction. While
single-view 3D reconstruction methods in the past have relied on expensive
supervised inputs such as part annotations or CAD models, our matching enables a
first approach towards a challenging new problem, namely, \new{part annotation
  free} single-view reconstruction. We discuss initial approaches to variants of existing supervised methods or structure from motion (SFM) pipelines that may be used to solve this problem without requiring annotations.

\vspace{-0.4cm}
\paragraph{Propagating correspondences}
In the CUB-200-2011 dataset, there are only 60 images for each
category. Moreover, birds are often imaged from preferred viewpoints,
but it is critical for reconstruction to obtain matches across a
well-distributed set of viewpoints. On the other hand, deformations may be very high even within a category (open wings as opposed to closed), which makes straightforward matching within a category challenging. Inspired by the work of Carreira \etal \cite{VVN}, we use a shortest path method to propagate matches across objects of similar shapes in the dataset, in order to obtain a denser set of tracks. However, note that we lack the initial set of point annotations as well as the camera poses obtained through part annotations in \cite{VVN,Vincente}, who also manually select a subset of keypoints to eliminate articulations. Instead, we determine unsupervised matches purely through our WarpNet and rely on the pose graph to determine nearest neighbors for propagation.

\vspace{-0.4cm}
\paragraph{Choosing a subset for reconstruction}
A key problem we encounter is the choice of images for reconstruction. In previous works on reconstruction within PASCAL VOC \cite{VVN,Vincente}, it has been possible to use the entire dataset since it contains less than 1000 images for birds. In contrast, CUB-200-2011 contains nearly 12000 images, which poses computational challenges and requires greater vigilance against outliers. Moreover, annotations in \cite{VVN,Vincente} preclude the need for algorithmic considerations on baseline or shape variations in choosing the image set. For instance, to reconstruct a sitting bird imaged from a frontal view, we must propagate matches to side views of sitting birds in other categories to ensure a good baseline, while avoiding images of flying birds. 

Given a collection of images, several heuristics have been proposed for selecting the right subset or order for multiview rigid-body reconstruction \cite{Snavely,SSS}. However, those are not directly applicable for single-view reconstruction of deformable objects. Instead, we propose three heuristics that utilize the structure of fine-grained bird datasets:
\begin{itemize}
\setlength\itemsep{-0.15cm}
\item Use images from categories that share a keyword (for example, all ``warblers'', or all ``sparrows'').
\item Use images from categories that are related by an ornithological taxonomy, as defined by \cite{BNA}.
\item Use images from the five nearest neighbor subcategories on a similarity tree of bird species \cite{Thomas}.
\vspace{-0.15cm}
\end{itemize}
The above heuristics perform comparably and address the same goal -- introduction of matched keypoints from more than one subcategory to ensure good viewpoint coverage.

\vspace{-0.4cm}
\paragraph{Reconstruction}
Given an image of a target object from one particular class, we consider images from several other categories using one of the above heuristics. We compute pairwise matches at $85\%$ precision threshold between all pairs of images whose distance on the pose graph is less than $4$. We ignore pairs that have less than $50$ surviving matches. We then set up a virtual view network \cite{VVN} to propagate matches across all the selected images.
We use scores from \eqref{eq:score}, bounded between $[0, 1]$, as
weights on the graphs connecting the keypoints. After propagation, we
discard as spurious any propagated matches with shortest path distance
more than $0.4$ and remove all images that have less than $30$ matches
with the target object. We then create the measurement matrix of
tracked keypoints of the target object. We only consider keypoints
visible in at least $10\%$ of the images as stable enough for
reconstruction. We finally send the observation matrix to the rigid
factorization method of \cite{Marques}, which robustly handles missing
data, to obtain 3D shape.\footnote{A rigid factorization suffices to
  produce good reconstructions since the dataset is large enough, but
  non-rigid methods alternately could be used.}

\section{Experiments}
\label{sec:experiments}
\vspace{-0.2cm}

We perform experiments on the CUB-200-2011 dataset which contains 11788 images
of 200 bird categories, with 15 parts annotated \cite{Wah}. We
reconstruct without part annotation, assuming 
objects are localized within a bounding box. We quantitatively
evaluate our matches using and extending the part annotations. \new{Next, we evaluate
the effectiveness of WarpNet as a spatial prior} and analyze the choice of transformations for creating the artificial training dataset. Finally, we demonstrate the efficacy of our framework with several examples of unsupervised single-view reconstruction.

\subsection{Experimental Details}
\label{exp}
\vspace{-0.2cm}
We create the pose graph of \cite{Krause} using the \texttt{conv4} feature of AlexNet trained on ILSVRC2012 \cite{Alex}. For creating the artificial dataset, we only use the training data ($\sim$6000 images) and create $m=9$ copies of each image using our exemplar-TPS. 
We resize all images to $224 \times 224$. 
This results in approximately 120k image pairs, each with $n = 100$ point correspondences. Following \cite{Dosovitskiy}, we apply spatial and chromatic data augmentation on-the-fly during training.

We use the VGG-M architecture of \cite{Chatfield} until the \texttt{pool5} layer as the feature extraction component of WarpNet. The point transformer consists of C512-C256-F1024-D-Op using the notation of \cite{Pulkit}. Both convolutional layers use 3x3 kernel, stride 1 with no padding, with ReLU non-linearity. The output layer is a regressor on the grid coordinates, with grid size $K=10$. The feature extraction weights are initialized with weights pre-trained on the ILSVRC classification task, following prior state-of-the-art for correspondence \cite{jlong}.

For matching and reconstruction, images are resized with aspect ratio intact and the smallest side $224$ pixels. We uniformly sample points on the foreground with a stride of 8 as keypoints for matching. For all experiments we use L2-normalized \texttt{conv4} features extracted at the keypoints using the hole algorithm \cite{Chen} for computing the appearance term in \eqref{eq:score}. Hyperparameters used for matching are $\sigma_f=1.75$, $\sigma_w=18$, $\lambda=0.3$, tuned using the artificial dataset.

\subsection{Match Evaluation}
\vspace{-0.2cm}
We compare our approach with ILSVRC pre-trained VGG-M \texttt{conv4} \cite{Chatfield}, SIFT at radius 8 \cite{lowe2004} and matches from the deformable spatial pyramid (DSP) \cite{Jaechul}.

Only the appearance term in \eqref{eq:score} is used for computing matches with
VGG-M \texttt{conv4} and SIFT. For computing the matches with DSP, we mask out
the background prior to extracting SIFT features following \cite{VVN} and only
keep matches of the keypoints. For this experiment, the set of keypoints to
match includes the locations of annotated parts.

\new{In order to evaluate WarpNet as a stand-alone learned spatial prior, we compare WarpNet with DSP by replacing the
SIFT features in DSP with VGG features. We call this method \emph{VGG+DSP}. We
further evaluate WarpNet against the original DSP by using WarpNet as a spatial
prior for SIFT matches, where the unary term $d_f$ in \eqref{eq:score} is computed with SIFT features. We call
this method \emph{SIFT+WarpNet}.}

\new{As discussed in Section \ref{sec:making}, the only supervision required in
  training WarpNet is the segmentation mask to mine exemplar-TPS
  transformations. We also evaluate the robustness of WarpNet using co-segmentation outputs of \cite{Krause}, called
  \emph{VGG+coseg}.}

\vspace{-0.4cm}
\paragraph{Test set}
We evaluate on 5000 image pairs that are within 3 nearest neighbors apart on the pose graph, comprising more than 50k ground truth matches. \footnote{Please see supplementary materials for results on a test set with 1-nearest neighbors, where we observe similar trends but with higher PCKs.}

Due to the unsupervised nature of the pose graph, these pairs exhibit significant articulation, viewpoint and appearance variations (see Figures \ref{fig:open}, \ref{fig:networkout}). 
We remove severely occluded pairs with less than 7 parts visible in both images and pairs whose TPS warp computed from part annotations have very high bending energy. None of the test images were used to train WarpNet.

\vspace{-0.4cm}
\paragraph{Evaluation metrics}
We evaluate the accuracy of matches with the percentage of correct keypoints (PCK) metric \cite{Yang}, where a match is considered correct if the predicted point is within $\alpha * L$ of the ground-truth correspondence.

Following \cite{Pulkit}, we chose $L$ to be the mean diagonal length of the two images. 
We also compute the precision-recall (PR) curve adopting the procedure of \cite{Mikolajczyk}. A match is considered a true positive within a radius $\alpha = 0.05$, otherwise it is a false positive. 
In this setup, a recall of 1 is obtained only if all the matches retrieved are correct, that is, $100\%$ $\alpha$-PCK. We compute PR curves using the ratio-test values described in Section \ref{sec:matchcost} for ranking the matches and report AP. For DSP, we use its matching cost for ranking instead of the ratios, since second closest matches are not available.

\begin{figure}[!!t]
\subfloat[]{\label{fig:pr1}\hspace{0em}\includegraphics[width=0.5\linewidth]{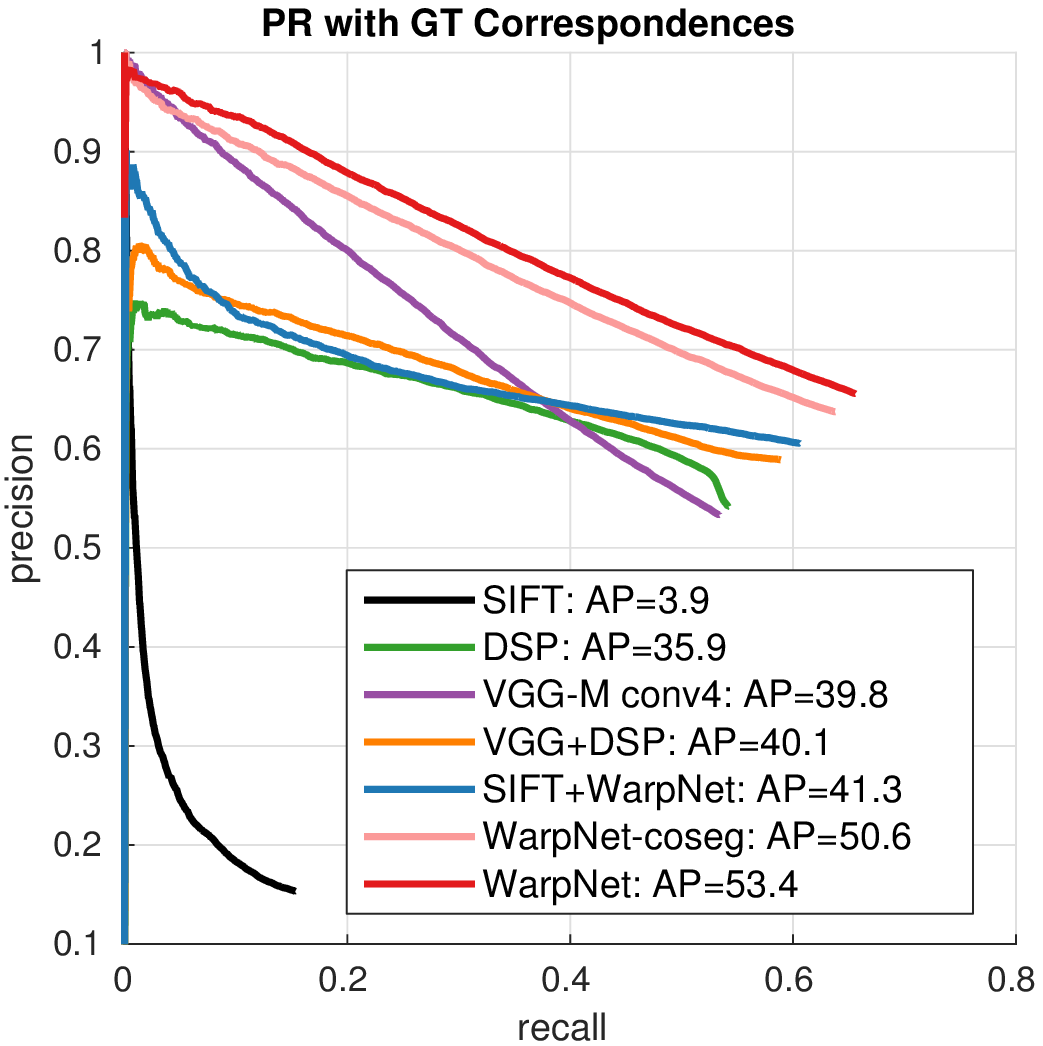}}
\subfloat[]{\label{fig:pr2}\includegraphics[width=0.5\linewidth]{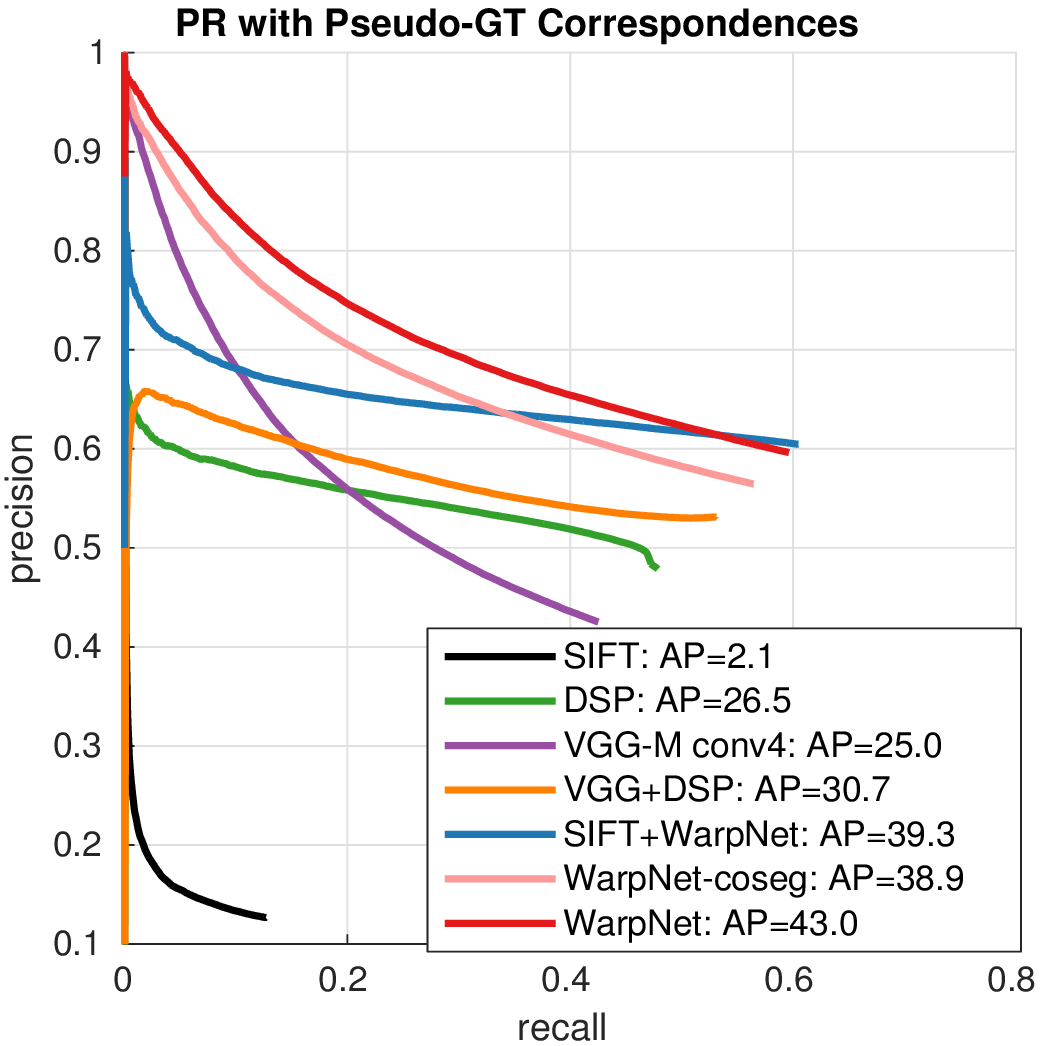}}
\vspace{-0.3cm}
\caption{\small Precision-Recall curves for matching points between neighboring images on the pose graph. We evaluate points with \protect\subref{fig:pr1} human-annotated correspondences and \protect\subref{fig:pr2} expanded pseudo-ground-truth correspondences.}
\label{fig:pr}
\vspace{-0.3cm}
\end{figure}

\vspace{-0.4cm}
\paragraph{Results}
Figure \ref{fig:pr}\subref{fig:pr1} shows the obtained PR curves. WarpNet
achieves an AP of $53.4\%$, an $13.6\%$ increase over matches using just the
appearance feature of VGG-M \texttt{conv4}. WarpNet achieves a much higher
recall due to its spatial prior, learned without using any part annotations. As
a side note, \texttt{conv4} features of WarpNet alone achieve very similar
performance to the VGG-M \texttt{conv4}. 
\new{In all cases, WarpNet outperforms DSP as a spatial prior and changing SIFT
  to VGG features yields around 5\% improvement in the final recall. \emph{WarpNet-coseg} still outperforms the baseline
  VGG-M by $10.8\%$, showing our approach is applicable even without ground truth segmentations. }

\begin{figure}[!!t]
\centering
\subfloat[]{\label{fig:pck1}\hspace{0em}
\includegraphics[width=0.45\linewidth]{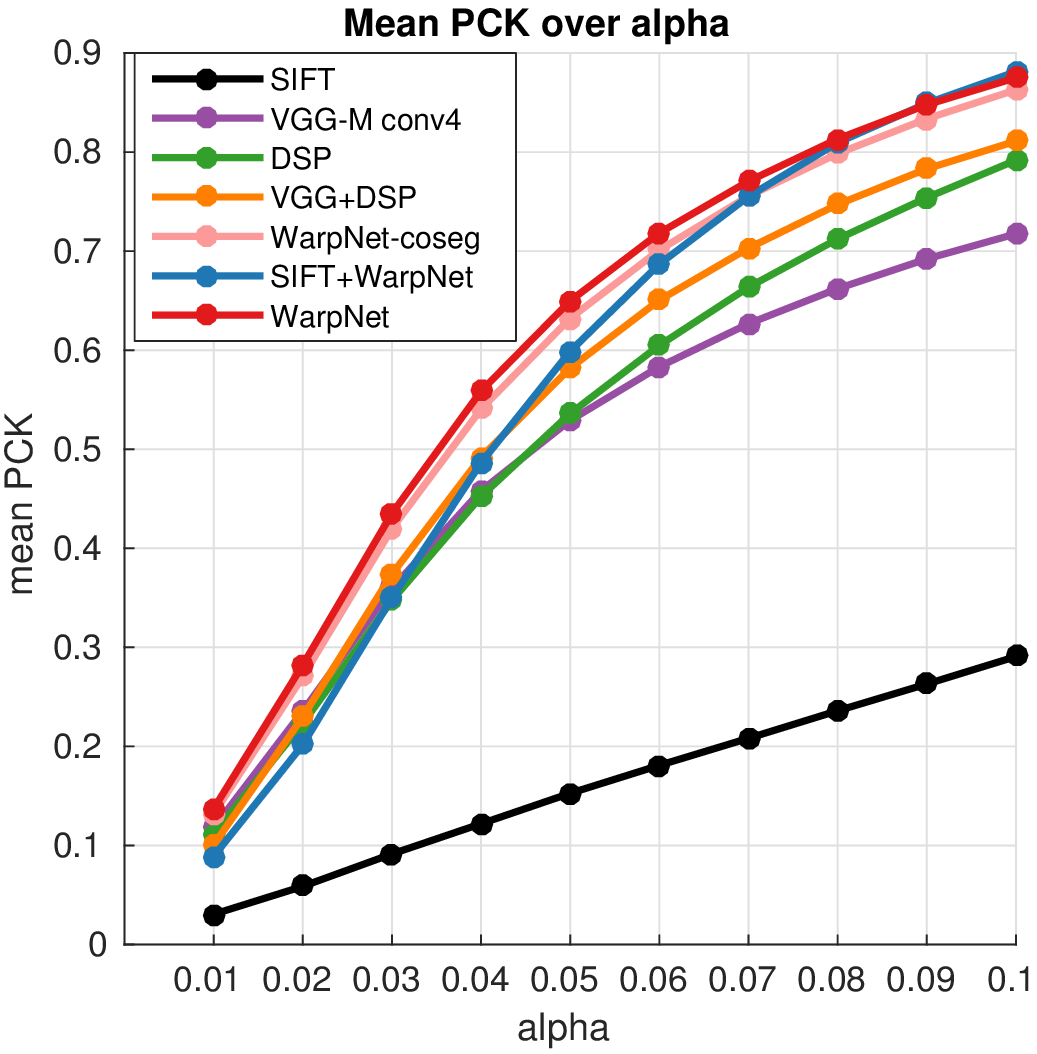}}
\subfloat[]{\label{fig:pck2}\includegraphics[width=0.45\linewidth]{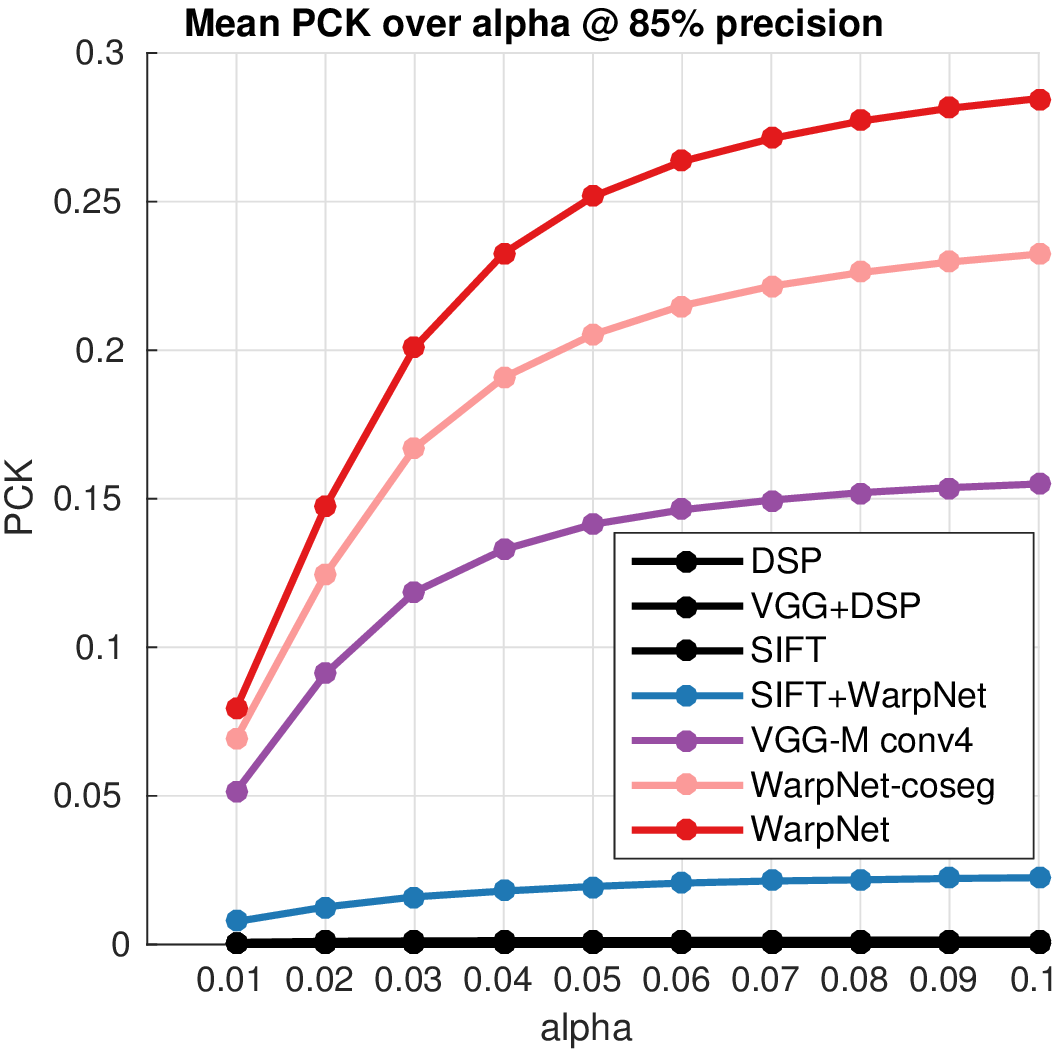}}
\vspace{-0.3cm}
\caption{\small PCK (higher the better) over varying definition of correctness $\alpha$. \protect\subref{fig:pck1} Mean PCK of all retrieved matches regardless of ratio score. \protect\subref{fig:pck2} Mean PCK with matches thresholded at $85\%$ precision, which are the matches used for reconstruction.}
\label{fig:pck}
\vspace{-0.3cm}
\end{figure}

Figure \ref{fig:pck}\subref{fig:pck1} shows the PCK as a function of $\alpha$,
where WarpNet consistently outperforms other methods. We observe that VGG-M
\texttt{conv4} and DSP perform similarly, showing that while \new{DSP obtains
  low recall at high precision, 
its overall} match quality is similar to CNN features, an observation in line with \cite{VVN}. Since only high precision matches are
useful for reconstruction where outliers need to be avoided, we show the same
curves thresholded at $85\%$ precision in Figure \ref{fig:pck}\subref{fig:pck2}
for VGG-M and our method. \new{Note that some methods in black have zero recall at this
precision.} The growing gap between WarpNet and VGG-M \texttt{conv4} as $\alpha$ increases suggests that, unlike WarpNet, appearance features alone make grossly wrong matches (see Figures \ref{fig:open} and \ref{fig:qual_match}).

\vspace{-0.4cm}
\paragraph{Expanding the set of part annotations}
\begin{figure}[!!t]
\begin{center}
\includegraphics[width=0.8\linewidth]{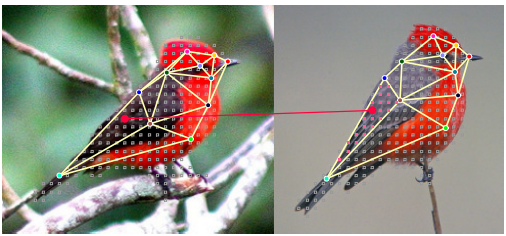}
\end{center}
\vspace{-0.5cm}
\caption{\small Illustration of the pseudo-gt correspondences. We triangulate each image using the annotated keypoints (colored points). The match for the big red dot in the left image is found by looking at points within the same triangle (small pink dots) in the right image and picking the closest point in terms of barycentric coordinates. 
}
\label{fig:pseudo}
\vspace{-0.3cm}
\end{figure}

A caveat of the CUB-200-2011 for our task is that part annotations are
sparse and concentrated on semantically distinct parts such as eyes
and beaks around the head region, with only four points on the bird
body that are often not all visible. To investigate matching
performance more densely, we carefully expand the ground-truth matches
using the annotated parts. This process is illustrated in Figure
\ref{fig:pseudo}. Given a pair of images $I_1$ and $I_2$, we Delaunay
triangulate each image independently using the parts visible in both
as vertices. For a point $u$ within a triangle in $I_1$, we consider
points in $I_2$ that are within the \emph{same} triangle as possible
candidates (shown as pink dots in Figure \ref{fig:pseudo}), find the
point that is closest to $u$ in terms of barycentric coordinates and
accept this as a new pseudo ground-truth match if the distance is less than $0.1$. Figure \ref{fig:pr}\subref{fig:pr2} shows the PR curve obtained using the pseudo-ground truth matches (in addition to the annotated parts). We see the same trends as Figure \ref{fig:pr}\subref{fig:pr1}, but with a wider gap between the baselines and our method. This is reasonable given that bird bodies usually consist of flat or repeated textures that are challenging to match with local appearances alone, highlighting the efficacy of WarpNet's spatial prior.

\subsection{Choice of Transformations}
\vspace{-0.2cm}

\begin{figure}
\centering
\subfloat[PR]{\label{fig:aff1}\hspace{0em}\includegraphics[width=0.45\linewidth]{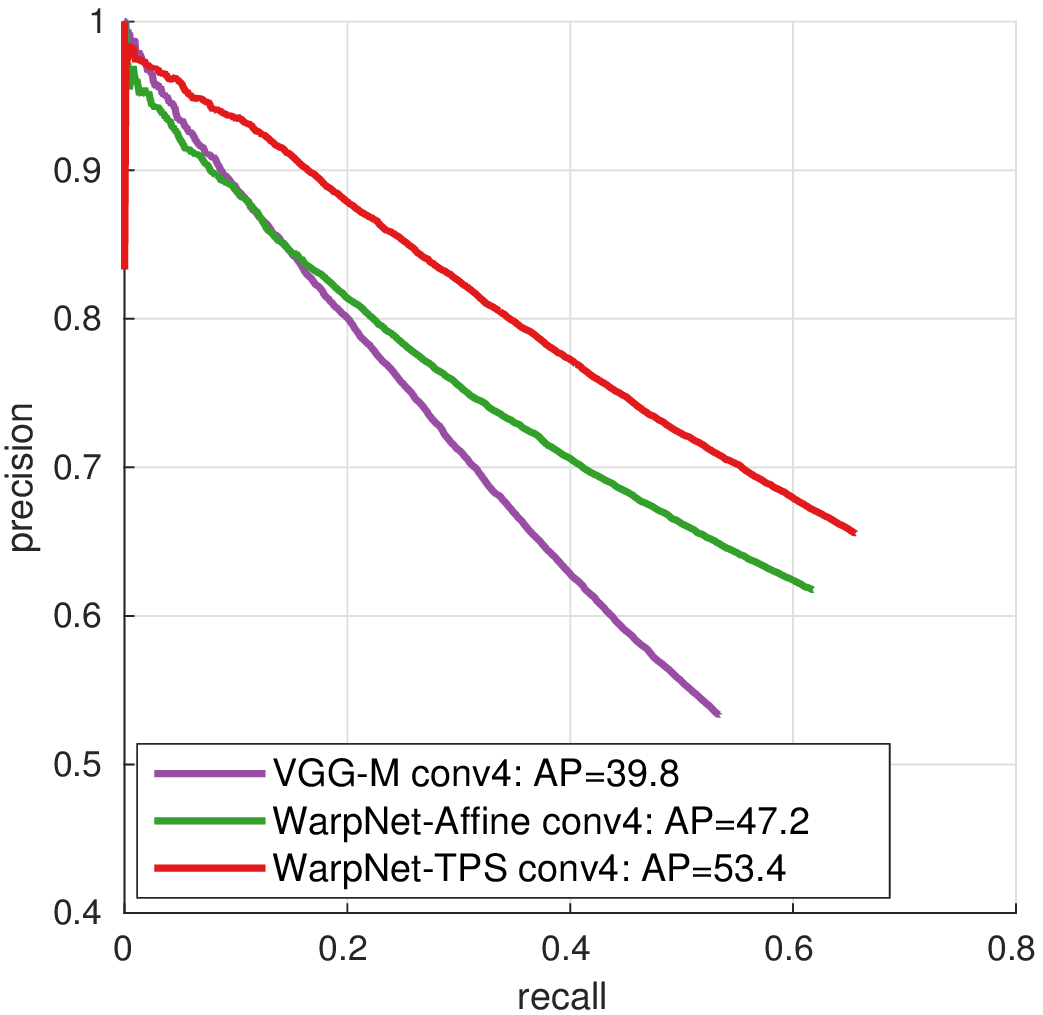}}
\subfloat[PCK@85\%]{\label{fig:aff2}\includegraphics[width=0.45\linewidth]{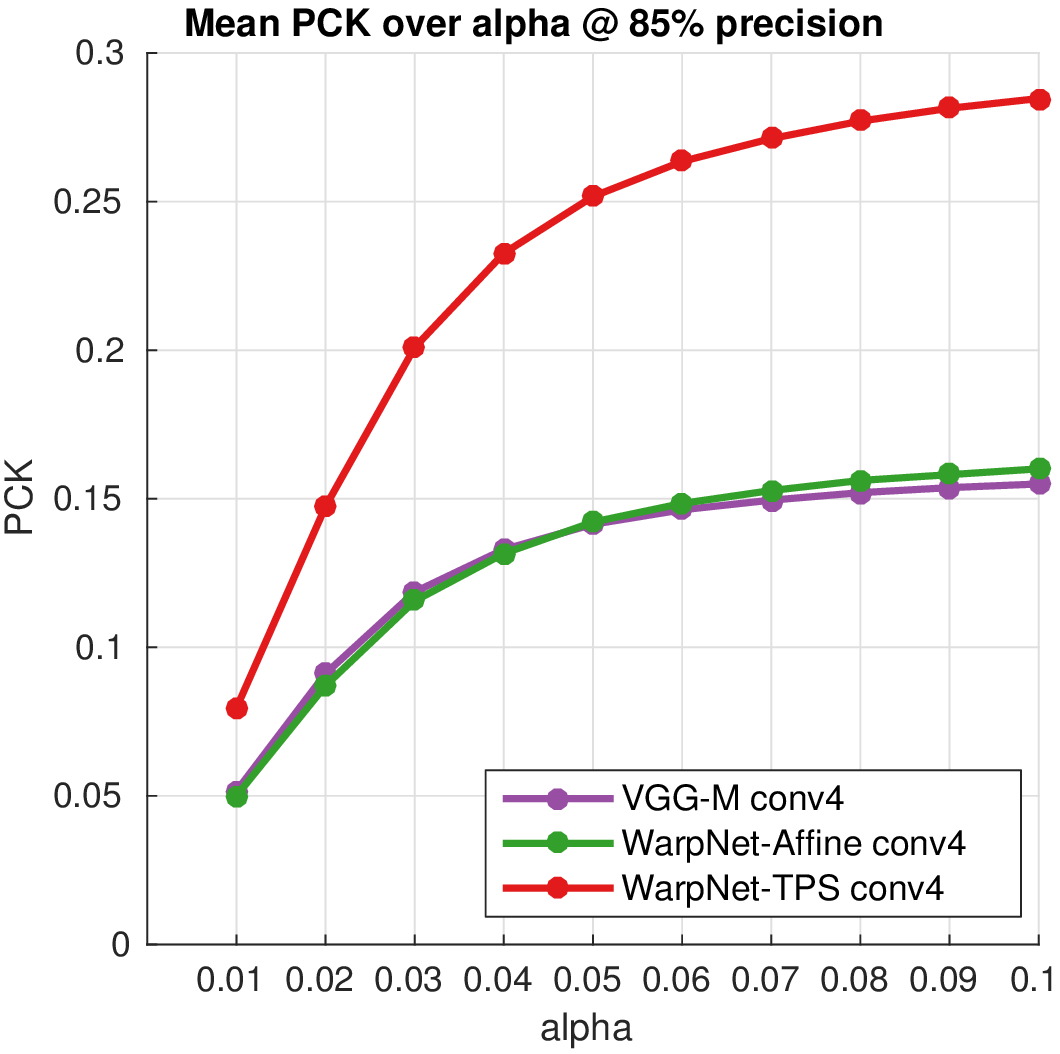}}
\vspace{-0.3cm}
\caption{\small Comparing results for WarpNet trained on artificial data created using affine-spatial transformations with \protect\subref{fig:aff1} PR curves and \protect\subref{fig:aff2} PCK over $\alpha$. WarpNet trained with exemplar-TPS is more effective in terms of recall and precision.}
\label{fig:affineNet}
\vspace{-0.4cm}
\end{figure}

We now analyze the choice of exemplar TPS transformations for creating the artificial dataset. We train another WarpNet under the same settings, but on an artificial dataset created using only affine spatial transformations, which we refer to as AffineNet. Note that AffineNet's output is still a TPS transformation, thus, it has the same capacity as the original WarpNet. Figure \ref{fig:affineNet}\subref{fig:aff1} shows the PR curve of AffineNet in comparison to WarpNet and VGG-M \texttt{conv4}. WarpNet outperforms AffineNet in all aspects. While AffineNet has a higher final recall (that is PCK of all matches) than VGG-M \texttt{conv4}, its recall at high precision is slightly lower than that of VGG-M \texttt{conv4}. This is highlighted in Figure \ref{fig:affineNet}\subref{fig:aff2}, which shows PCK of matches at $85\%$ precision over $\alpha$, where AffineNet performs on par with VGG-M \texttt{conv4}. This indicates that the warps predicted by AffineNet are helpful in a general sense, but not precise enough to improve the recall at high precision. This experiment shows that using exemplar-TPS transformations for creating the artificial dataset is critical for training a useful WarpNet.

\begin{figure*}
\begin{center}
\vspace{-0.1cm}
\includegraphics[width=0.93\linewidth]{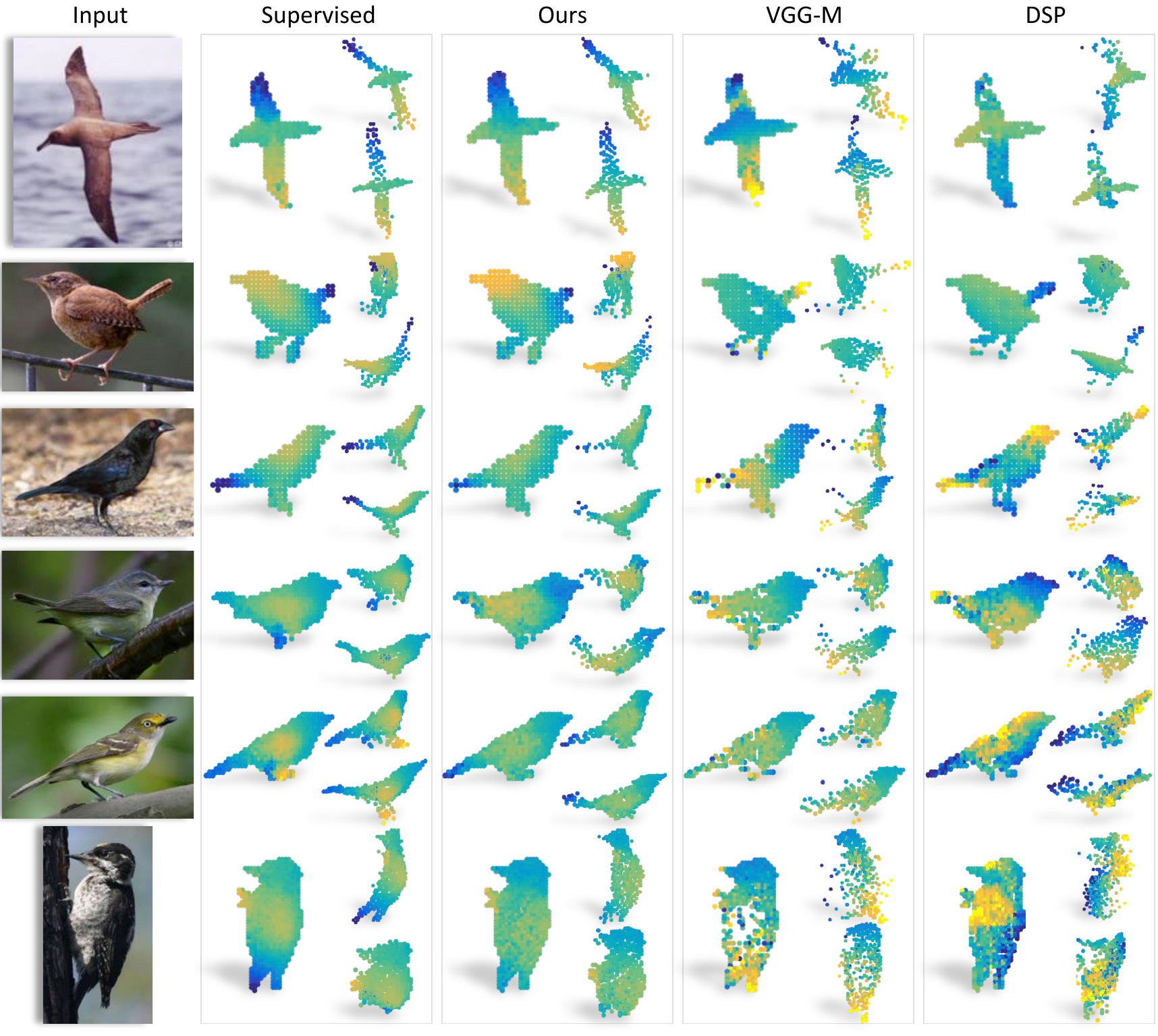}
\end{center}
\vspace{-0.5cm}
\caption{\small Sample reconstructions showing 3 views for each
  method: The camera viewpoint followed by the $45^{\circ}$ azimuth in
  counter-clockwise direction (top right) and $45^{\circ}$ elevation
  (bottom right). Colors show the depth where yellow is closer and
  blue is farther. The supervised method uses the spatial prior computed from annotated part correspondences, which can be seen as an upper bound. No part correspondences were used for the last three methods. WarpNet consistently obtains reconstructions most similar to the supervised method.}
\label{fig:reconst}
\vspace{-0.4cm}
\end{figure*}

\subsection{Single-view Object Reconstruction}
\vspace{-0.2cm}

We compare our method with three other matching methods. One is a supervised
matching approach similar to \cite{VVN}, where the network predicted TPS warp
$T_\theta$ in \eqref{eq:warpcost} is replaced by the supervised TPS warp
computed using the annotated keypoints. We call this approach
\texttt{supervised} and it is an upper-bound to our method since ground-truth
part annotations are used for reconstruction. We also perform reconstructions
with VGG-M \texttt{conv4} features alone and DSP. We do not include the mirrored
image as another viewpoint of the target object, since bilateral symmetry does
not hold for articulated objects. For post-processing we use the xy-snapping method proposed in \cite{VVN}, which only uses the $z$-component from the reconstructed shape, while fixing the $x$, $y$ coordinates. We do not resample the target objects multiple times prior to factorization since it did not seem to make a difference.

Figure \ref{fig:reconst} shows reconstructions for various types of birds using the four methods from three viewpoints: camera view, $45^{\circ}$ azimuth and $45^{\circ}$ elevation. The colors indicate depth values (yellow is close, blue is far), with range fixed across all methods. WarpNet produces reconstructions that are most consistent with the \texttt{supervised} approach. Reconstructions from VGG-M and DSP are noisy due to errors in matching and often produce extreme outlier points that had to be clipped for ease of visualization. Articulated parts such as tails and wings are particularly challenging to match, where VGG-M and DSP often fail to recover consistent depths. 
A weakness of our method is that the TPS prior may sometimes hallucinate birds of similar pose even with wide baseline. This may be avoided by better choice of images for reconstruction. Please see supplementary material for more results, qualitative matches and reconstruction videos.

\vspace{-0.2cm}
\section{Conclusions and Future Work}
\vspace{-0.2cm}
We introduce a framework for matching and reconstruction in fine-grained datasets that avoids the expense and scalability challenges of part annotations. 
The core of our approach is a novel deep learning \new{architecture} 
 that predicts a function to warp one object into another. We show that our network can be trained without supervised part annotations by exploiting the structure of fine-grained datasets and use its output as a spatial prior for accurate matching.
Our approach achieves significant improvements over prior state-of-the-art without using part annotations and we show reconstructions of similar quality as supervised methods. Key challenges for future work are to determine optimal subsets of images for reconstruction and a good order for adding images that allows incremental reconstruction with bundle adjustment.

\vspace{-0.2cm}
\paragraph{Acknowledgments}
This work was part of A.~Kanazawa's internship at NEC Labs America, in
Cupertino. A.~Kanazawa and D.~Jacobs were also supported by the National Science
Foundation under Grant No. 1526234.

{\small
\bibliographystyle{ieee}
\bibliography{paper}
}

\end{document}